\RequirePackage{amsmath}
\documentclass[runningheads]{llncs}
\usepackage{lmodern}
\usepackage[utf8]{inputenc}
\usepackage{latexsym}
\usepackage[T1]{fontenc}
\usepackage{amssymb}
\usepackage{stmaryrd}
\usepackage{eulervm}
\usepackage{palatino}
\usepackage{mdwmath}
\usepackage{mdwtab}

\usepackage{graphicx}
\usepackage[cmyk]{xcolor}
\usepackage{tikz}
\usepackage{tabularx}
\usepackage{array}
\usepackage{eqparbox}
\usepackage{booktabs}

\usepackage{amssymb}
\usepackage{makecell}
\usepackage{diagbox}

\usepackage{array}
\usepackage{lineno}

\usepackage{caption}
\usepackage{url}
\usepackage{hyperref}
\usepackage{enumitem}

\newcommand{\pc}{\mathbf{P}}

\newcommand{\bsi}{\boldsymbol{\beth}}

\usepackage{enumitem}

\title{Representing Pedagogic Content Knowledge Through Rough Sets}
\titlerunning{Content Knowledge Through Rough Sets}
\author{\textsf{A Mani}\thanks{This research is supported by Woman Scientist Grant No. WOS-A/PM-22/2019 of the Department of Science and Technology.}}
\authorrunning{A Mani}
\institute{Machine Intelligence Unit, Indian Statistical Institute, Kolkata\\
203, B. T. Road, Kolkata-700108, India\\
Email: \texttt{$a.mani.cms@gmail.com$} \texttt{$amani.rough@isical.ac.in$}\\
Homepage: \url{https://www.logicamani.in}\\
Orcid: \url{https://orcid.org/0000-0002-0880-1035} }

\begin{document}

\maketitle

\begin{abstract}
A teacher's knowledge base consists of knowledge of mathematics content, knowledge of student epistemology, and pedagogical knowledge. It has severe implications on the understanding of student's knowledge of content, and the learning context in general. The necessity to formalize the different content knowledge in approximate senses is recognized in the education research literature. A related problem is that of coherent formalizability. Existing responsive or smart AI-based software systems do not concern themselves with meaning, and trained ones are replete with their own issues. In the present research, many issues in modeling teachers' understanding of content are identified, and a two-tier rough set-based model is proposed by the present author for the purpose of developing software that can aid the varied tasks of a teacher. The main advantage of the proposed approach is in its ability to coherently handle vagueness, granularity and multi-modality. An extended example to equational reasoning is used to demonstrate these. The paper is meant for rough set researchers intending to build logical models or develop meaning-aware AI-software to aid teachers, and education research experts.

\end{abstract}


\section{Introduction}

A number of granular and nongranular semantics of rough sets are known in the literature. Concepts of knowledge can be associated with these from multiple perspectives such as those based on classical rough ideas \cite{zpsk07}, mereological axiomatic granular perspectives \cite{am5586,am501,am9969}, classical granular computing \cite{ya01}, interpretations of modal logic \cite{ppm2,amedit}, constructive logic \cite{jpr}, concept analysis \cite{yy2016,cd9,am9969}, evidence theory, and machine learning. These ideas of knowledge are not uniformly well-developed across these types of rough sets. Possible applications of these to education research, and specifically mathematical education research is a very domain-specific matter. Many aspects of these are explored by the present author in her earlier papers \cite{am5559,am23r,am9114,am1113,am2022c}.

Application of rough sets and other artificial intelligence methodologies to education research is important for all the fields involved. This is because of the intense complexity of knowledge representation in education research. The subject is substantially population specific, and is strongly influenced by socioeconomic, psychological, and cultural factors. The very idea of \emph{core content} for a level of learning as understood from the perspective of a subject specialist is not reasonably definable without additional constraints. The field of marginal mathematics is, for example, handled through very different approaches (see for example, \cite{am2354}). However, the idea of core content in mathematics (for example) for regular courses in middle school or higher can be specified though not exactly as one would want to. 

In the modern view \cite{bdtmp2008}, three components that determine a teacher's knowledge base are knowledge of mathematics content, knowledge of student epistemology, and pedagogical knowledge. The first is about the breadth and depth of a teacher's knowledge of the subject. The second concerns understanding of the psychological principles of learning, concept formation, assimilation, and maturation. Teacher's understanding of teaching in the light of the other components is knowledge of pedagogy. In \cite{ghkl2004}, for example, a very instructive application of the concepts is presented in brief, and it is shown how a teacher's fixation with some computational procedures for understanding functions can adversely affect conceptual learning outcomes of the class. In the example, the teacher chose not to use the second of the problems suggested (as \textit{"it is about chemistry"}): 
\begin{enumerate}
 \item {A pharmacist is to prepare 15 ml of $2\%$ eye drops for a glaucoma patient. However, she has only $10\%$ and $1\%$ solutions in stock. Can she compound these to fill the prescription?}
 \item {Help her find a convenient way of dispensing different volumes of $2\%$ eye drops for other patients using the same solutions in stock.}
\end{enumerate}
Further associations between a teacher's beliefs about the nature of the subject (such as \textsf{school mathematics is a fixed set of concepts and procedures that are to be delivered to and remembered by students}), and the fixations are deducible. These aspects matter for both the training of, and designing intelligent software aids for teachers (and additionally for the development of curricula and textbooks). 

A student's understanding of a mathematics textbook in relation to the instructional context is distinct from a teacher's understanding of the pedagogic content (including psychological aspects) and context  -- the latter, in turn typically differs from a subject expert's view on the matter (see \cite{bdtmp2008} for a detailed discussion). The categories of students, teachers and experts may be classifiable into a number of finer categories on the basis of subsets of features. The classification of conceptual knowledge as evidenced by the work produced by learners (using any form of expression or in any settings or context) takes many forms. While concept inventories are useful to access over many types of contexts, they do not typically have the breadth to consider atypical reasoning, and related errors. \emph{Teachers understanding of the pedagogic content is in relation to all these, and specifically to the meta-level language used for the context}. The purpose of this research to propose a framework for this idealized understanding.  In actual practice, teachers may themselves have their own misunderstandings, unfounded beliefs and biases, and may lack suitable training for the job. The last aspect is ignored here, and therefore the research is about an idealized understanding. Teachers and students, in a classroom, actually approximate a body of knowledge formed in a distributed cognition perspective \cite{am24e} in different ways.

The real numbers are taught in certain languages with much interspersed vagueness in middle school. Further, mathematical problem-solving methods, and concepts have vagueness in related discourse. Making use of predicates such as \textsf{is an approximation of}, \textsf{is a superset of}, \textsf{is a subset of}, and \textsf{is definitely a part of} much of the formalization is possible. The harder problem of expressing them through general rough approximations is additionally explored in this research after formalizing the school real numbers typically ignored by formal mathematicians. Work in the area is primarily limited because of the need to systematically confront the varieties of vagueness inherent in it \cite{jsr2022,icmi2021}. However, there is much work on pragmatic approaches to the ontology of the real numbers in the context. They serve as sources for building viable models.

In the context of mathematics education research, the very idea of \emph{formal language or model} is open for some debate. In \cite{am2022c}, it is argued that mereology combined with a language of approximations can potentially be used to build higher order formalizations of concepts that go beyond the restrictions envisaged in \cite{jsr2022} or in earlier work \cite{icmi2021}. 

Further, a common practice in teaching mathematics is to use everyday language when describing and explaining ideas. Teachers may use phrases such as \emph{plug in} a value to evaluate a function, and \emph{cancel} instead of dividing by the common factor. These and the overuse of pronouns can obscure the meaning of the procedures and concepts being used. While learning from errors can be a productive exercise \cite{rbor96}, the use of imprecise, loose language, and unidentified errors by teachers can be counterproductive \cite{acasvm19}. These and related studies motivate the need to build reasonably formal models to study  teachers' knowledge of content, and more so for the purpose of building intelligent aids for teachers. ML approaches based on numeric simplifications of language features or keywords cannot succeed in the context, and none are attempted in the literature. 

The authors of \cite{jsr2022}, specify the concept of \emph{coherent formalizability} thus: \textit{"By coherence, we mean that formalisation of disparate elements hangs together as a meaningful whole. The classroom discourse includes many parts pertaining to definitions, visualisations, representations, conjectures, exemplification, providing counterexamples, justification, and  refutation".}
Formal models are however not offered in the mentioned paper. This problem is addressed in more detail here through the machinery of partial algebraic systems, mereology, and general rough sets.

To summarize, the following are done in this research (within the domain of pedagogic knowledge):
\begin{itemize}
\item {A much improved applicable model of the vague version of real numbers is proposed;}
\item {The model is intended for use with a companion model that involves multiple general rough approximation operators, mereological and rough predicates (possibly functional), and }
\item {a justification of aspects of the above is argued for through equational reasoning contexts. }
\end{itemize}
Directions for future research are additionally provided.

\section{Background}

Partial operations, and partial algebraic systems pervade this research. For the basics of partial algebras, the reader is referred to \cite{lj}. A \emph{partial algebra} $P$ is a tuple of the form \[\left\langle\underline{P},\,f_{1},\,f_{2},\,\ldots ,\, f_{n}, (r_{1},\,\ldots ,\,r_{n} )\right\rangle\] with $\underline{P}$ being a set, $f_{i}$'s being partial function symbols of arity (or place-value) $r_{i}$. The interpretation of $f_{i}$ on the set $\underline{P}$ should be denoted by $f_{i}^{\underline{P}}$; however, the superscript will be dropped in this paper as the application contexts are simple enough. If predicate symbols enter into the signature, then $P$ is termed a \emph{partial algebraic system}.   

In this paragraph the terms are not interpreted. For two terms $s,\,t$, $s\,\stackrel{\omega}{=}\,t$ shall mean, if both sides are defined then the two terms are equal (the quantification is implicit). $\stackrel{\omega}{=}$ is the same as the existence equality (sometimes written as $\stackrel{e}{=}$) in the present paper. $s\,\stackrel{\omega ^*}{=}\,t$ shall mean if either side is defined, then the other is and the two sides are equal (the quantification is implicit). Note that the latter equality can be defined in terms of the former as 
\[(s\,\stackrel{\omega}{=}\,s \, \longrightarrow \, s\,\stackrel{\omega}{=} t)\&\,(t\,\stackrel{\omega}{=}\,t \, \longrightarrow \, s\,\stackrel{\omega}{=} t) \]

A partial weak lattice is a partial algebra of the form $L =\left\langle \underline{L},\vee,\wedge  \right\rangle$ (with $\underline{L}$ being a set) that satisfies
\begin{align*}
a\wedge a =a = a\vee a. \; (a\wedge b)\wedge c \stackrel{\omega}{=} a\wedge (b\wedge c). \tag{wpl1}\\
(a\vee b)\vee c \stackrel{\omega}{=} a\vee (b\vee c) \tag{wpl2}\\
(a\wedge b)\vee a \stackrel{\omega}{=} a. \; a\wedge b \stackrel{\omega}{=} b\wedge a. \;  a\vee b \stackrel{\omega}{=} b\vee a \tag{wpl3}
\end{align*}

\subsection{Rough Concept Inventories}\label{cibasic}

A test that focuses on evaluating a student's competence in a specific skill is a criterion-referenced test. Usually a person's test scores are intended to suggest a general statement about their capabilities and behavior. Concept inventories (CIs) are criterion-referenced test designed to test a student's functional understanding of concepts. However, they are mostly used by education researchers to assess the effectiveness of pedagogical methods. Computer-based assessment software do use conceptual models such as labeled conceptual graphs and formal concept analysis. But related exercises  require careful formalism to avoid misunderstanding and automatic evaluation is known to miss conceptual problems \cite{prn2013}. In the present author's view this is also because they try to avoid (rather than confront) vagueness inherent to the available knowledge.

Concept inventories are not well-suited for student-centered methods of evaluation because they make use of questions in the multiple-choice query format (MCQ) alone. The reason for choice of incorrect or correct answers can be elicited through additional requirements of explanations \cite{kmvs15}. In \cite{am5559}, \emph{rough concept inventories} are proposed by the present author. These are intended as modification of the methods of a concept inventory for effectively handling vagueness inherent in relatively student centric perspectives. Rough sets are used to represent approximate evaluations of explanations. However, the quality of the methodology is bound to depend on that of the underlying language used, and the latter is explored in this research. 

\paragraph{Mereology:}
Mereology \cite{ham2017} consists of a number of theoretical and philosophical approaches to relations of parthood (or \emph{is a part of} predicates) and relatable ones such as those of \emph{being connected to, being apart from, and being disconnected from}. Such relations can be found everywhere, and they relate to ontological features of any body of soft or hard knowledge (and their representation).  Mereology is used within general rough sets by the present author \cite{am9969,am501,am5586,am240} and others \cite{rjle2007,pls}. The subject is applied to education research in \cite{am2022c,am5559,am2021ff,am23r} by her. 

\subsection{Negations and Implications}
Some essential generalized implications and negations (for more details, see for example \cite{am9915,bsmm2022}) on a bounded partially ordered set with top element $\top$ and bottom $\bot$ are mentioned here. 
Consider the conditions possibly satisfied by a map $n: L\longmapsto L$:
\begin{align}
n(\bot) = \top \, \&\, n(\top) = \bot \tag{N1}\\
(\forall a, b) (a\leq b \longrightarrow n(b)\leq n(a)) \tag{N2}\\
(\forall a) n(n(a)) = a \tag{N3}\\
n(a) \in \{\bot , \top \} \text{ if and only if } a= \bot \text{ or } a = \top \tag{N4}
\end{align}

$n$ is a \emph{negation} if and only if it satisfies \textsf{N1} and \textsf{N2}, while $n$ is a \emph{strong negation} if and only if it satisfies all the four conditions.

Implications satisfy a wide array of properties as they depend on the other permitted operations. Here some relevant ones are mentioned. 

A function $\bsi : L^2 \longmapsto L $ is an \emph{implication} if it satisfies (for any $a, b, c \in L$) the following:
\begin{align}
\text{If } a \leq b \text{ then } \bsi bc \leq \bsi ac \tag{First Place Antitonicity FPA}\\
\text{If } b \leq c \text{ then } \bsi ab \leq \bsi ac \tag{Second Place Monotonicity SPM}\\
\bsi \bot\bot =\top \tag{Boundary Condition 1: BC1}\\
\bsi \top \top = \top \tag{Boundary Condition 2: BC2}\\
\bsi \top \bot = \bot  \tag{Boundary Condition 3: BC3}
\end{align}

Some other properties of interest in this paper are 
\begin{align*}
\bsi \top x = x \tag{LNP}\\
\bsi a(\bsi bc) = \bsi (b \bsi (ac)) \tag{Exchange Principle EP}\\
\bsi ab =1 \text{ if and only if } a\leq b   \tag{Ordering Property, OP}\\
\bsi a(\bsi ab) = \bsi ab  \tag{Iterative Boolean Law, IBL}\\
b \leq \bsi ab \tag{Consequent Boundary, CB}\\
\bsi aa = \top \tag{Identity Principle, IP}\\
\bsi a(\bsi bc) = \bsi (\bsi ab)(\bsi ac) \tag{T3}\\
\bsi (\bsi ab)b = \bsi (\bsi ba)a \tag{T4}
\end{align*}

\section{The Real Numbers and the Formalizability Problem}

In \cite{am2021ff,am23r}, formal structures associated with the course content of middle school students was not explored as the primary concern was about coherently introducing negative numbers to primary school students. Here the partial algebraic systems are enlarged to fit the larger context, and for the purposes of relating it to teachers knowledge of content. 

From the algebraic perspective, primary and middle school students typically learn subsets of the real numbers over a permissive implicit model with a very long signature. No distinction is made between operation symbols and their interpretations (interpreted operations), the symbol $+$ is interpreted both as a binary and a unary operation. The same is the case with the symbol $-$, while $a\div b$ may also be written as $\frac{a}{b}$. For reference, omitting exponentiation, the signature is \[\Sigma = (\leq, \geq, +, \times , \div,  -, \frac{}{}, \oplus, \ominus, \surd, 0, 1, (2, 2, 2, 2, 2, 2, 2, 1, 1, 1, 0, 0) .\] 
The numbers in the braces refer to the place value of operation symbols, and the \emph{plus and minus signs} have been denoted as $\oplus$ and $\ominus$ respectively. \emph{The standard conception of terms in universal algebra is partially split into a number of concepts such as multiplicative terms, binomials, monomials, rational fractions, and fractions in school algebra}. Multiplicative terms are also simply referred to as terms, and even these are only implicitly provided with a recursive definition. The concept of \emph{variable symbols} and their interpretation over domains is standard in universal algebra. However, the concept of a \emph{variable} is used in a loose way without a clean specification of the domains in school textbooks. 

The predicates $\leq$ and $\geq$ are used in the sense of \emph{being numerically less than} or \emph{being numerically greater than}. In the context of word problems, ideas of some objects being bigger or smaller than others often leads to ambiguity and vagueness. This is especially true when objects under consideration have multiple attributes associated. It is also the case that transitive parthood is frequently associated with the $\leq$ relation. Functional non-transitive parthood is apparently not commonly used in mathematics text books at least -- though this typically happens in the context of activities and descriptions that are very contextual and under-specified. For example, $3$ is the greatest prime less than $5$ and $5$ is the greatest prime less than $7$. However, $3$ is not the greatest prime less than $7$. All this means that the addition of an additional parthood predicate $\pc$ to the signature is a good idea. However, the question remains that \emph{Where should such predicates be used} (since multiple models are intended for the procedure)?  Such parthood predicates should be kept in the companion model to avoid complicating the concept of vague reals further. In \cite{am23r}, this aspect is not clarified. 
 
Relative to the former signature, the partial algebraic systems that are taught to middle school children are related to $\mathbb{R} = \left\langle\underline{X}^*, \,\Sigma  \right\rangle$ with $\underline{X}^*$ being a subset of the set corresponding to the algebraic closure of the union of algebraic numbers and $\{\pi\}$. For convenience, this can simply be taken as the algebraic numbers because the transcendental nature of $\pi$ is never taught before high school. Some properties satisfied by the partial algebraic system are as follows:

\begin{align*}
(\forall a, b) (a\leq b \leftrightarrow b\geq a)\tag{def1}\\
+,\, \times \text{ are weakly commutative, and associative} \tag{asc1}\\
(\forall a, b, c)\, a\times (b+c) = (a\times b) + (a\times c)  \tag{l-dist}\\
(\forall a, b, c)\,  (b+c)\times a = (b\times a) + (c\times a)  \tag{r-dist}\\
(\forall a)\, a + 0 = a\times 1 = a  \tag{identity}\\
(\forall a) \ominus \ominus a = a = (0-(0- a)) \tag{minus1}\\
(\forall a) a + \ominus a = 0 = a - a \tag{minus2}\\
(\forall a) \oplus a = a = 0+ a = a + (\oplus a)  \tag{plus1}\\
(\forall a, b) (a\leq b \leftrightarrow \ominus b\leq \ominus a)\tag{sgnord}\\
(\forall a, b) (a\leq b \leftrightarrow b\geq a)\tag{def1}\\
(\forall a, b, c) (a\leq b \longrightarrow a+c \leq b+c)\tag{comp1}\\
(\forall a, b) (a\leq b \leftrightarrow 0\leq b - a )\tag{ordiff}\\
(\forall a, b) a\leq b \text{ or } b\leq a \tag{totalord}\\
(\forall a, b) (\ominus a) \times (\ominus b) = a \times b \tag{sign1}
\end{align*}
\begin{align*}
(\forall a, b) (\ominus a) \times (\oplus b) = \ominus (a \times b) \tag{sign2}\\
(\forall a, b) (\ominus a) \times (\ominus b) = a \times b \tag{sign1}\\
(\forall a) a+ (\ominus a) = 0 \tag{sign3}\\
(\forall a, b) (b\neq 0 \longrightarrow (a \div b)\times b = a )   \tag{div1}\\
(\forall a, b, c) (b\neq 0 \longrightarrow (a \div b)\times c = (a\times c)\div b )   \tag{div2}\\
(\forall a, b) (a+ b = 0 \leftrightarrow b = \ominus a) \tag{inv+}\\
(\forall a ) a\div 0 = \text{ undefined} \tag{div3}\\
(\surd a)^2 \stackrel{\omega}{=} a \tag{su1}\\
\surd a \surd b \stackrel{\omega}{=} \surd (ab) \tag{su2}
\end{align*}
The above system will be referred to as the \emph{Vague Real Number System}. 

It is not that every student understands or assumes all the axioms mentioned above simultaneously. It can be expected that these are partially ordered in distinct ways in the context of problem-solving, and moreover 
\begin{itemize}
\item {Only certain subsets ($X$ say) of these and variants may be understood as the \emph{basic axioms},}
\item {from the basic axioms, only certain subsets of derivations may be used, and }
\item {in addition alternate conceptual axioms (possibly higher order) may be assumed.}
\end{itemize}
All this can be used to explain only aspects of diverse problem-solving behaviors, and concepts of negative numbers adhered to. 
Further, linguistic diversity (due to natural languages, and artificial modes of communication) has far-reaching implications on reasoning about the axioms. For a recent overview, the reader is referred to \cite{icmi2021}. Alternate conceptual axioms can, for example, arise from the use of different types of negation in a language. Intuitionist negations that work with adjectives, for example, are not uncommon. Such differences may have a long history as in Chinese mathematics \cite{am2022c}. Apart from the colored rod systems with their own computational processes associated, in the context of astronomical computations, concepts of directional strong and weak approximations of real numbers were considered as \emph{positive} and \emph{negative} (If a larger number approximates a given number it is a stronger or positive approximation, while approximations that are smaller are \emph{negative}. 

As of now, explicit mathematical models of the reals are not used often enough in the associated discourse at both pedagogic content knowledge and subject levels. However, the need to incorporate \emph{subclasses of such partial algebraic systems} for relatively more exact discourse is compelling. More so because of the broader developments in artificial intelligence.

\section{Extended Example: Equational Reasoning}

In middle school, skills in equational logic (in a loose sense), the ability to solve simultaneous and linear equations in at most two variables, and related word problems are an essential part of the course contents. These can be related to implicit or explicit concept inventories associated, and several distinct methodologies may be adopted to solve problems in the domain. Examples of weak equational logic that can potentially be used (even in school algebra) may be found in \cite{wc1989}. This is because substitution rules form hierarchies of their own, and partial interpretation of operations are not uncommon. Further, the rigor aspect can be restricted to specific types of operations on the real numbers, and related formulas.

\begin{figure}[hbt]
\begin{center}
\begin{tikzpicture}[node distance=2cm, auto]
\node (A) {Co};
\node (B) [above of=A] {MP};
\node (CO) [right of=B] {MS};
\node (C) [right of=A] {TC};
\node (E) [below of=A] {T};
\node (EO) [left of=A] {C};
\node (F) [below of=EO] {SC};
\draw[<->,font=\scriptsize,thick] (A) to node {}(B);
\draw[<->,font=\scriptsize,thick] (A) to node {}(C);
\draw[<->,font=\scriptsize,thick] (B) to node {}(CO);
\draw[<->,font=\scriptsize,thick] (A) to node {}(E);
\draw[<->,font=\scriptsize,thick] (C) to node {}(CO);
\draw[<->,font=\scriptsize,thick] (EO) to node {}(F);
\draw[<->,font=\scriptsize,thick] (EO) to node {}(E);
\draw[<->,font=\scriptsize,thick] (CO) to node {}(A);
\draw[<->,font=\scriptsize,thick] (F) to node {}(E);
\draw[<->,font=\scriptsize,thick] (E) to node {}(C);
\end{tikzpicture}
\caption{Components of the Ecosystem}\label{latt}
\end{center}
\end{figure}
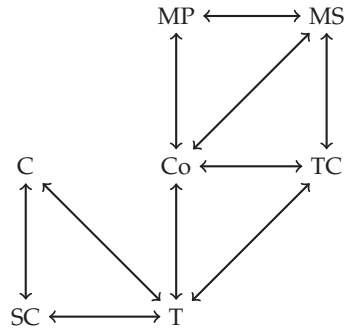

Every other middle school mathematics textbook specifies the sub-concepts (such as \emph{the operation of adding or subtracting both sides of an equation by a constant or variable does not alter its solution}) that a student is expected to be familiar with. Let a representation of the students' knowledge of content be \textsf{SC}. and the objective content be $C$. Now, an ideal teacher of the subject can be required to know a lot more about $C$, the learning of \textsf{SC} by a student of type $\tau \in \mathcal{T}$ (where $\mathcal{T}$ is a categorization of students by their sociocultural context) \textsf{TC}, and the relation of \textsf{SC} to the teacher's understanding of the content \textsf{T}, and relevant concepts \textsf{Co}. Partial models \textsf{MP} of \textsf{Co}, and sufficient models \textsf{MS} of the ideal teachers' knowledge of content \textsf{T} may be constructed from the information. Thus, the components of the ecosystem may be represented by the interrelationship diagram Fig \ref{latt} (the hierarchies are left to the reader to understand).

\subsection{How to Specify Approximations?}
 
Many concepts generated by students are upper approximated by a concept in the real or virtual teacher's knowledge of content. It is very much possible that relatively unrelated concepts of students are approximated by the same upper approximation. This is not very helpful in practice, as it points to the use of insufficiently fine-grained granules in the process of constructing approximations, and additionally, using an insufficient number of approximation operations. Optimizing over these choices is important, and is illustrated below. Optionally, approximations can be constructed using the methods of \cite{am24e} by treating the steps of proofs as granules, adding additional granules, and specifying an up-directed \emph{possibly contains} relation. 

Suppose that a number of students solve the problem $\text{Solve } 2x+3 = 4x+1$. Let six different refined versions of their solutions be as below. Such a scenario may be the case when the problem is offered in a student-centric mode of learning in a larger context. A non-student centric learning algebra context might yield some of the first four responses. Starred steps refer to failed conceptions or attempts at comprehending rules. Each of the solutions can be characterized relative to the concept formation, assimilation, and maturity of the nearest proper solution of the type demonstrated. For example, Solution-5 is a proper version of Solution-1. Part of \textsf{Solution-1} is correct, and a \emph{much greater part of} \textsf{Solution-2} is correct. While predicates such as \emph{much greater part of} might appear to be a higher order fuzzy predicate, it would be better to characterize them with upper approximations. Writing $(Solution-1)^{uuu} = Solution-5 = (Solution-2)^{uu}$ is certainly a meaningful interpretation of a non-idempotent abstract upper approximation. $u$ annihilates negative or insufficient or defective understanding of specific operations or rules, and is therefore a special kind of upper approximation. 

\textsf{Solution-1}
\begin{description}
 \item [Step-1]{$2x+3 = 4x+1$ (Given)}
 \item [Step-2]{$x+3 = 2x+1$ ($*$ Cancelling 2 from both sides).}
 \item [Step-3]{$x+3-3= 2x+1 -3$ (Subtracting 3 from both sides)}
 \item [Step-4]{$x= 2x-2$ (Simplifying)}
 \item [Step-5]{$x-x= 2x-2-x$ (Subtracting $x$ from both sides)}
 \item [Step-6]{$0=2-x $ and so $x=2$ (Transposing)}
\end{description}

\textsf{Solution-2}
\begin{description}
 \item [Step-1]{$2x+3 = 4x+1$ (Given)}
 \item [Step-2]{$2x+3 -1 = 4x$ (Transposing $1$}
 \item [Step-3]{$2x+2= 4x$ (Simplifying)}
 \item [Step-4]{$x+2= 2x$ ($*$ Cancelling the factor $2$)}
 \item [Step-5]{$x+2-x= 2x-x$ (Subtracting $x$ from both sides)}
 \item [Step-6]{$x=2 $ ($*$ Transposing)}
\end{description}
The steps of a solution may be seen as concept abstractions, and approximated. Step-2 of \textsf{Solution-1} and Step-4 of \textsf{Solution-2} apparently share a similar kind of conceptual error. However, the level of misconception in the former is more than in the latter.  Therefore, it is safe to assert that a negation of the latter is part of a negation of the former. 

\textsf{Solution-3}
\begin{description}
 \item [Step-1]{$2x+3 = 4x+1$ (Given)}
 \item [Step-3]{$2x+3-1= 4x+1 -1$ (Subtracting $1$ from both sides)}
 \item [Step-4]{$2x+2= 4x$ (Simplifying)}
 \item [Step-5]{$2x+2-2x= 4x-2x$ (Subtracting $2x$ from both sides)}
 \item [Step-6]{$2=2x$ (Simplifying)}
 \item [Step-7]{$x=1$ (Cancelling 2 from both sides).}
\end{description}

\textsf{Solution-5}
\begin{description}
 \item [Step-1]{$2x+3 = 4x+1$ (Given)}
 \item [Step-2]{$0= 4x+1 -(2x+3)$ (Transposing $2x+3$ to RHS)}
 \item [Step-3]{$0= 2x-2$ (Simplifying)}
 \item [Step-4]{$2=2x$ (Transposing $2$ to LHS))}
 \item [Step-5]{$x=1$ (Cancelling the common factor $2$ from both sides).}
\end{description}
  
\textsf{Solution-6}
\begin{description}
 \item [Step-1]{$2x+3 = 4x+1$ (Given)}
 \item [Step-2]{Let, $f(x)$ be the function defined by $f(x)= 4x+1 -(2x+3)$ )}
 \item [Step-3]{The graph of the function is a straight line that intersects the $X$-axes at $x=1$ only, and so it is the required solution of the equation.}
\end{description} 
 
\textsf{Solution-9}
\begin{description}
 \item [Step-1]{$2x+3 = 4x+1$ (Given)}
 \item [Step-2]{Let, $f(x)$ be the function defined by $f(x)= 4x+1 -(2x+3)$ )}
 \item [Step-3]{$f(0) =-2 <0$, while $f(2) =2 > 0$.}
 \item [Step-4]{A linear equation can have zero, one or infinitely many solutions. }
 \item [Step-5]{The function changes sign between $x=0$ and $x=2$. So its solution is between $0$ and $2$.}
 \item [Step-6]{The graph of the function from $x=0$ to $x=2$ is a straight line segment that intersects the $X$-axes at $x=1$, and so it is the required solution of the equation.}
\end{description} 

Any comparison of \textsf{Solution-9 or 6} with the rest requires additional approximate predicates that works across subdomains. Specifically, the comparison of the quality of \textsf{Solution-6} and \textsf{Solution-5} can be done through an inventory of common features. This is suggestive of functional parthood predicates \cite{ham2017} such as \emph{is a substantial part of} \cite{am9015} or \emph{is at least as good as} among many others. While the retention of such predicates is useful, it is always better to replace them with ones that are functional (mathematically). More importantly, most teachers would not want to mix up graphical and equational reasoning. Such a desire can be manifested through multiple approximation operators (possibly partial).

\section{The Companion Algebraic Rough Systems}

Multiple systems may be proposed that fit the constraints specified, and those implicit in the previous sections. It can be seen that the partial algebraic system should have enough order structure, parthood predicates (or other mereological predicates), multiple total or partial lower and upper approximation operators, meaningful negations and implications at least. In addition, granulations in the axiomatic sense \cite{am5586} are desirable as they are closely tied with meaning (though the requirement that all abstract approximations be granular would be too strong for all use cases within the context). In addition to these, a predicate \textsf{is an approximation of} $\mathsf{A}$ is permitted, though its use is strongly discouraged.

In \cite{am9915}, a minimalist model for rough sets called a rough convenience lattice (\textsf{RCL}) is introduced, and it is shown that it is equivalent to RCL aggregation negation algebras, and RCL aggregation implication algebras in a perspective. However, a minimalist model such as those of a rough convenience lattice (RCL) \cite{am9915} or its well-justified abstract generalizations \textsf{CRCLANA} or \textsf{CRCLAIA} may not be optimal to build upon as the lattice order structure cannot be expected to be shared among the many approximation operators that are desired. A bounded quasi-ordered set would be more appropriate for the purpose. The elements of the base set are intended to correspond to sets of statements about the reals or related structures. 

\begin{definition}\label{rcon}
An partial algebraic system of the form ${B} \, =\, \left\langle \underline{B}, l, u, \leq, \vee,  \wedge, \bot, \top \right\rangle$ with $(\underline{B}, \leq, \bot, \top )$ being a bounded quasi-ordered set will be said to be a \emph{rough convenience quasi-order} (RCQO) if the following conditions are additionally satisfied ($\vee$ and $\wedge$ being partial weak lattice operations, and the operations $l$ and $u$ are generalized lower and upper approximation operators respectively):
\begin{align*}
(\forall a, b) a\vee b= b \text{ or } a\wedge b= a \longrightarrow  a\leq b \tag{wl12}\\
(\forall a, b, c) (a\vee b= c \text{ or } c\wedge b= a \longrightarrow a\leq c) \tag{wl34}\\
(\forall x) x^{ll}= x^l \leq x \leq x^u \leq x^{uu}  \tag{qlu1}\\
(\forall a, b) (a\leq b \longrightarrow a^l\leq b^l ). \; (\forall a, b) (a\leq b \longrightarrow a^u\leq b^u )    \tag{qlu-mo}\\
a^u \vee b^u \stackrel{\omega}{=} (a\vee b)^u. \; (a\wedge b)^l \stackrel{\omega}{=} a^l \wedge b^l   \tag{qlu23}\\
\top^u = \top \,\&\, \bot^{l} = \bot =\bot^u  \tag{topbot}
\end{align*}
\end{definition}

\begin{definition}\label{arclaia}
An \emph{Abstract RCQO Aggregation Implication Algebra} (RQOAI) is an algebra of the form ${B} \, =\, \left\langle \underline{B}, \leq, \otimes, \cdot, \vee,  \wedge, l, u, \bsi_\neg, \bsi_\sim \bot, \top \right\rangle$ that satisfies:
\begin{align*}
\left\langle \underline{B}, \vee,  \wedge, l, u, \bot, \top \right\rangle \text{ is a RCQO.}  \tag{rcl}\\
(a^{uu} = a^u \& b^{uu} = b^u \&  e^{uu} = e^u \longrightarrow a\otimes (b \otimes e)= (a\otimes b)\otimes e) \tag{wAasso1}\\
a\otimes((b\vee e)\otimes a) \stackrel{\omega}{=} ((a\vee b )\otimes c)\otimes c  \tag{wAsso2}\\
\bsi_\sim \text{ satisfies FPA, SPM, BC3, and IBL}.   \tag{imsc}\\
\bsi_\neg \text{ satisfies FPA, IP, SPM, BC1, BC2, and BC3}.   \tag{inegc}
\end{align*}
with $\cdot$ being a commutative, monoidal, order-compatible operation with identity $\bot$, and $\otimes$ additionally being a commutative, order-compatible operation with identity $\top$.
\end{definition}

These considerations lead to the following proposed model over the signature (for a finite integer $n>0$) of type $(2,2,2,2,1,1,1,\ldots, 1, 1, \ldots, 1,0, 0)$ \[\Sigma = (\pc, \leq, \otimes, \cdot, \bsi_\neg, \bsi_\sim, l_1, \ldots l_n, u_1, \ldots, u_n, \bot, \top ).\]

\begin{definition}
An \emph{ER Companion Model} shall be a partial algebraic system of the form $S= \left\langle\underline{S}, \Sigma\right\rangle$ (with $\underline{S}$ being a set) that satisfies 
$\left\langle \underline{B}, \leq, \otimes, \cdot, \vee,  \wedge, l_i, u_i, \bsi_\neg, \bsi_\sim \bot, \top \right\rangle$ is a \textsf{RCQO} for each $i\in \{1, 2, \ldots n\}$, and \textsf{UL1, UL2, UL3, and TB} hold \textsf{for some} $i\in \{1, 2, \ldots n\}$, and the others hold universally.
\begin{align*}
(\forall x) \pc xx ; \: (\forall x, b) (\pc xb \, \&\, \pc bx \longrightarrow x = b) \tag{PT1, PT2}\\
(\forall a, b) a\vee b \stackrel{\omega}{=} b\vee a  ; \;  (\forall a, b) a\wedge b \stackrel{\omega}{=} b\wedge a \tag{G1}\\
(\forall a, b) (a\vee b) \wedge a \stackrel{\omega}{=} a  ; \;  (\forall a, b) (a\wedge b) \vee a \stackrel{\omega}{=} a \tag{G2}\\
(\forall a, b, c) (a\wedge b) \vee c \stackrel{\omega}{=} (a\vee c) \wedge (b\vee c) \tag{G3}\\
(\forall a, b, c) (a\vee b) \wedge c \stackrel{\omega}{=} (a\wedge c) \vee  (b\wedge c) \tag{G4}\\
(\forall a, b) (a\leq b \leftrightarrow a\vee b = b \,\leftrightarrow\, a\wedge b = a  ) \tag{G5}\\
(\forall a \in \mathbb{S})\,  \pc a^l  a\,\&\,a^{ll}\, =\,a^l \,\&\, \pc a^{u}  a^{uu}  \tag{UL1}\\
(\forall a, b \in \mathbb{S}) (\pc a b \longrightarrow \pc a^l b^l \,\&\,\pc a^u  b^u) \tag{UL2}\\
\bot^l\, =\, \bot \,\&\, \bot^u\, =\, \bot \,\&\, \pc \top^{l} \top \,\&\,  \pc \top^{u} \top  \tag{UL3}\\
(\forall a \in \mathbb{S})\, \pc \bot a \,\&\, \pc a \top    \tag{TB}
\end{align*}
\end{definition}

Granularity is often clearly identifiable in education research contexts. The existing axiomatic frameworks for granularity such as those of \emph{high general granular operator space} \cite{am5586,am501} can be directly adapted for the purpose in the light of the assumptions on an ER companion model. The usual approach would be to regard basic concepts as granules.

\section{Directions}

In this research, a formal approach to the problem of reasonably formalizing teachers' knowledge in middle school mathematics is proposed. 
The proposed system consists of three major blocks: an improved better suited description of the real numbers as understood in school mathematics, formal systems for reasoning about mathematical assertions that accommodate mereology and vagueness, and psychosocial factors, and a granular extension of the same based on earlier work of the present author. Student-centric rough concept inventories \cite{am5559} can additionally be handled more smoothly through the companion models introduced as automatic evaluation of explanations is dependent on the latter. The coherent formalizability issue in the education research literature is thus addressed to an extent. The suggested higher-order formalizations of equational reasoning (including \cite{am24e}) are not unique, and making these more unique is an important problem. 

The purpose of the proposal is to characterize the underlying ontology, improve understanding, and see the impact of many interventions at a formal level. Additionally, it should be usable in focused contexts such as the study of the study of equational reasoning or reasoning about fractions or soft geometry through Geogebra, for a deeper understanding of their potential, and limitations. Future computational models for the contexts, if at all possible, may be assessed through the frameworks introduced.  Issues relating to use of multiple approximation operators, ideas of overlap predicates, generalized overlap functions, weak rough implications, rationality, and substantial parthood \cite{am9015}, will be investigated further in future work. The extent of decision-making afforded by the companion systems analogous to those using ortho-pairs \cite{sbbg2020} is additionally of interest. Additionally, this research motivates higher order approaches in rough sets, that would have been previously glossed over through hybrid layers.

\bibliographystyle{splncs04.bst}
\bibliography{algrough23c.bib}

\end{document}